# A benchmark multimodal oro-dental dataset for large vision-language models

Haoxin Lv[1], Ijazul Haq[2,3,*], Jin Du[4], Jiaxin Ma[5,6], Binnian Zhu[5], Xiaobing Dang[4], Chaoan Liang[7], Ruxu Du[4], Yingjie Zhang[2], Muhammad Saqib[8]

**Abstract:** The advancement of artificial intelligence in oral healthcare relies on the availability of large-scale multimodal datasets that capture the complexity of clinical practice. In this paper, we present a comprehensive multimodal dataset, comprising 8775 dental checkups from 4800 patients collected over eight years (2018-2025), with patients ranging from 10 to 90 years of age. The dataset includes 50000 intraoral images, 8056 radiographs, and detailed textual records, including diagnoses, treatment plans, and follow-up notes. The data were collected under standard ethical guidelines and annotated for benchmarking. To demonstrate its utility, we fine-tuned state-of-the-art large vision-language models, Qwen-VL 3B and 7B, and evaluated them on two tasks: classification of six oro-dental anomalies and generation of complete diagnostic reports from multimodal inputs. We compared the fine-tuned models with their base counterparts and GPT-4o. The fine-tuned models achieved substantial gains over these baselines, validating the dataset and underscoring its effectiveness in advancing AI-driven oro-dental healthcare solutions. The dataset is publicly available, providing an essential resource for future research in AI dentistry.

## 1 Background & Summary

The application of AI in dentistry has seen significant growth due to AI's capacity to enhance diagnostic accuracy, streamline processes, and improve patient care. Recent advancements in AI have facilitated unprecedented improvements in areas such as automated disease detection [1-3], predictive analytics [4], and personalized patient care within dental practices [5]. While most studies on AI in dentistry have primarily focused on traditional machine and deep learning approaches [6-10], research incorporating advanced transformer-based models such as Vision-Language Models (VLMs) and Large Multimodal Models (LMMs) remains scarce. Traditionally, deep learning methods require highly structured and annotated datasets, imposing substantial constraints on data type and format [10, 11]. Although traditional deep learning models perform well on specific datasets, their generalization abilities significantly deteriorate when encountering novel data. For example, a model trained to classify certain dental conditions often struggles when presented with previously unseen or inadequately represented anomalies [10].

In contrast, LMMs have intrinsic generalization capabilities, possessing elements reminiscent of Artificial General Intelligence (AGI), enabling them to adapt to tasks and recognize categories not explicitly seen during training [12]. LMMs are notably resilient to variations in input data [13]; an LMM trained on multiple modalities, such as images, X-rays, and textual data, can reasonably infer and provide decisions even when provided with incomplete information. This characteristic is particularly advantageous in situations where all modalities may not

Haoxin Lv and Ijazul Haq contributed equally.

[1] Department of Oral Implantology, Suzhou Doctor Dental Clinic Co. Ltd, Suzhou, 215000, China.
[2] Shien-Ming Wu School of Intelligent Manufacturing, South China University of Technology, Guangzhou, 511442, China.
[3] Artificial Intelligence Department, Guangdong CAS Angels Biotechnology Co. Ltd, Foshan, 528200, China.
[4] Guangdong Janus Biotechnology Co. Ltd, Guangzhou, 511400, China.
[5] Zhejiang CAS Angels Biotechnology Co. Ltd, Zhejiang, 314200, China.
[6] OMRON SINIC X Corp., Tokyo, 113-0033, Japan.
[7] MingZheng Dental Clinic, Guangzhou, 511400, China.
[8] Department of Software Engineering, University of Engineering & Technology, Peshawar, 25000, Pakistan.
* Corresponding Author: Email address (hanjie@sjtu.edu.cn)

always be simultaneously available. Consequently, an AI dental assistant leveraging an LMM can offer meaningful assistance regardless of the completeness of the input data, significantly improving accuracy as the data becomes more comprehensive. Moreover, LMMs impose fewer restrictions on input formats, effectively handling data in diverse forms including PDFs, textual notes, webpages, JPEG, PNG, and other commonly used file formats. This flexibility substantially reduces the resources and time typically required for data preprocessing in conventional deep learning pipelines.

Despite recent progress, a critical review of related works reveals a clear shortage of publicly accessible datasets tailored for training dentistry-specific VLMs. Summarized in Table 1, existing multimodal resources remain limited in scope. Some contain only radiographs, such as Panoramic X-rays (PaX), Periapical X-rays (PeX), and Cone-Beam Computed Tomography (CBCT) scans [14, 15], while others combine radiographs with textual reports [16]. The TDD dataset [17], for example, includes radiographs annotated with eye-tracking maps and audio from think-aloud protocols, but it was not designed for VLMs, and current transformer architectures are not optimized for such modalities. The MMDental [18] dataset is more comparable, integrating radiographs, CBCT scans, and photographs. However, it lacks textual diagnostic records, essential for VLM tasks, and remains relatively small in scale. These shortcomings highlight the need for a large, comprehensive multimodal dataset to advance generative AI in dentistry.

To address these limitations and foster progress in intelligent dentistry, we introduce **COde** (**C**asangels **O**ro-**de**ntal), a comprehensive multimodal dataset. COde contains 8775 dental checkups from 4800 patients collected over eight years (2018–2025), including 50000 intraoral RGB images, 8056 radiographs, and detailed patient records with treatment notes, follow-up reports, and medical histories spanning up to 8 years. Its scale and diversity exceed existing datasets, setting new standards for benchmarking dental AI. Covering patients aged 10 to 90 years and both genders, COde offers critical insights into oro-dental disease prevalence and progression. By releasing COde publicly, we aim to accelerate multimodal AI research in dentistry.

| Dataset | # of patients | Modalities (size) |
|---|---|---|
| Huang et.al [14] | 169 | PaX (8); PeX (16203); CBCT (329) |
| Wang et.al [15] | ~ 4000 | PaX (4000); CBCT (148400) |
| Silva et.al [16] | 1050 | PaX (8795); Text Reports (8029) |
| TDD [17] | 1000 | PaX, Audio, Gaze Map (1000 pairs) |
| MMDental [18] | 660 | PaX (2540); PeX (1120); CBCT-3D (430); Photos (3200) |
| **COde (Ours)** | 4800 (8775 checkups) | PaX, PeX, CBCT-2D (8056); Photos (50000); Text Reports (8775) |

Table 1. Summary of multimodal dental datasets, comparing prior works with the proposed COde dataset.

## 2 Methods

### 2.1. Data Collection

An overview of the dataset development process is illustrated in Fig. 1. Data were collected from routine patient visits at the Suzhou Dental Doctor Outpatient Clinic. Each check-up included multiple intraoral photographs, corresponding radiographic images, and a detailed textual clinical record. Intraoral photographs were taken prior to any treatment or intervention using a Canon D60 DSLR camera with a 100 mm macro lens, operated by assistant dentists, with patients seated in the dental chair under standard overhead lighting. PeX and PaX radiographs were obtained using standard dental X-ray units commonly employed in clinical practice. CBCT scans were performed by certified radiology technicians using a Sirona Galileos CBCT system. Each check-up also included a comprehensive diagnostic report written by the attending dental doctor in text form.

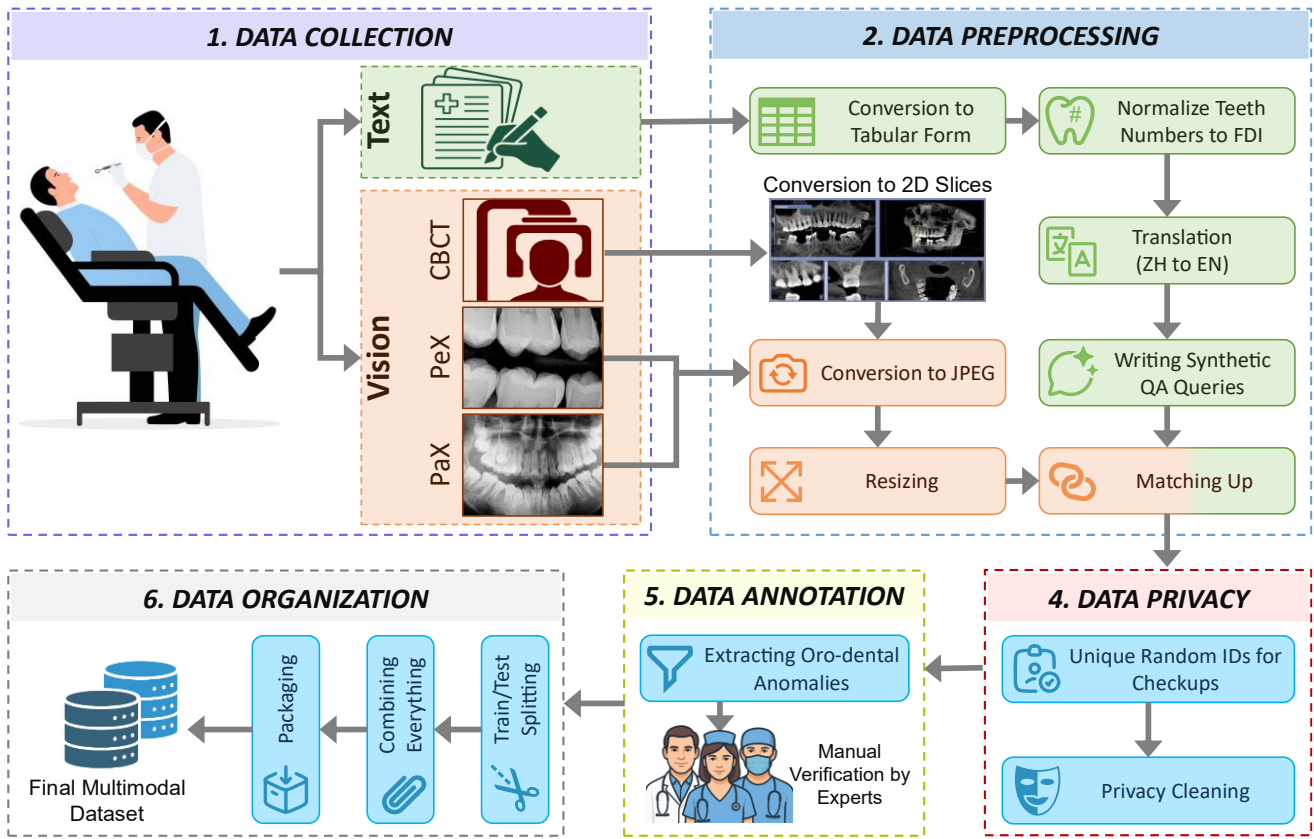

Fig. 1 Illustration of the dataset development methodology

All intraoral images, radiographs, CBCT scans, and accompanying patient records were managed using a cloud-based dental practice management system, E-KanYa[9], to securely store patient data and electronic records. All data were collected as part of standard care and subsequently archived for research use under proper ethical approvals.

**2.2. Data Preprocessing**

The textual portion of the dataset originates from the clinic's electronic medical records. Duplicate or irrelevant images were removed, and incomplete records lacking either text notes or images were excluded. Teeth numbers were originally in Palmer notation and were normalized to the FDI system. All images were converted to JPEG format and resized so that the longer dimension was set to 448px while preserving the original aspect ratio. All reports were originally written in Chinese by dentists as routine clinical records, they were translated into English using GPT-4o, resulting in a bilingual dataset where every field is available in both languages. While translation was not required for model training or fine-tuning, it enhances usability for researchers unfamiliar with Chinese. The original Chinese text is preserved alongside its English version for fidelity. Key fields recorded are listed as follows, and a random translated sample from the data is given in Fig. 2

– *Checkup ID:* Unique identifier for each dental check-up
– *Checkup Time:* Date and time when the check-up took place
– *Patient ID:* An anonymized unique identifier for the patient, links multiple check-ups of the same patient.
– *Age:* Patient's age (in years) at the time of the visit
– *Gender:* Patient's gender (recorded as male or female)
– *Photographs:* File names of intraoral photographs taken during the visit and stored in JPEG format

---

[9] E-KanYa: https://www.linkedcare.cn/kouqiang/about-eky

– *Radiographs:* File names of dental radiographic images (CBCT slices, PeX, and PaX) stored in JPEG format
– *Patient Record:* Main Clinical note summarizing the visit, including consultation, diagnosis, and treatment
– *Chief Complaint:* Primary issue or concern prompting the visit, as reported by the patient or guardian
– *Present Illness:* Description of the current dental issue's history and symptoms
– *Past Medical Record:* Relevant past medical/dental history (e.g., prior conditions, treatments, allergies)
– *Examination:* Key findings from the clinical examination of the oral cavity (e.g., condition of teeth and gums)
– *Radiographs Examination:* Key findings from the radiographic examination
– *Diagnosis:* Oro-dental condition(s) or diagnosis identified during the visit (often with standard codes)
– *Treatment Plan:* Planned treatments or procedures to address the diagnosed conditions
– *Management:* Details of the actual treatments or procedures performed during the visit
– *Medical Instructions:* Post-treatment care instructions and guidance given to the patient
– *Remarks:* Any additional notes or comments by the dentist (e.g., special considerations or administrative notes)

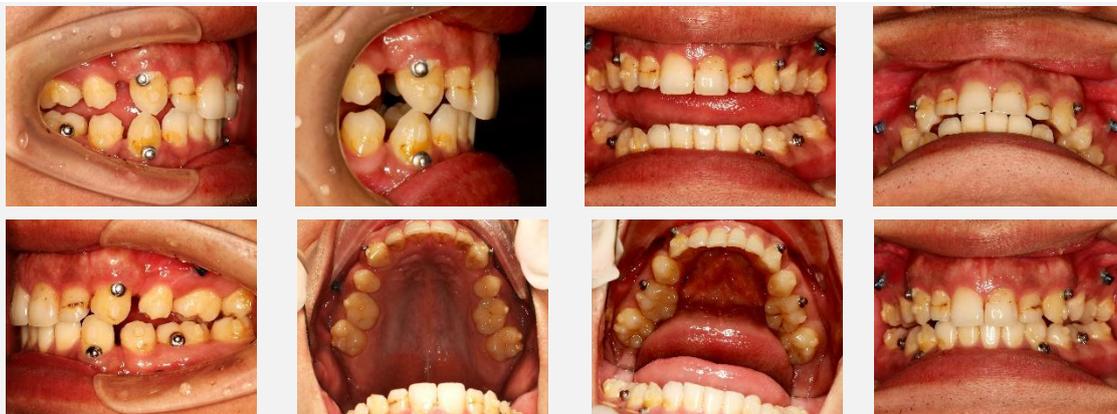

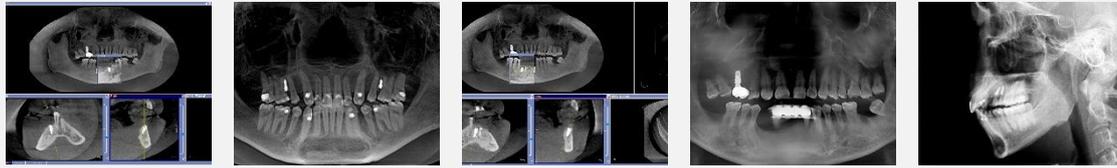

**Photographs**

**Radiographs**

**Diagnostic Report**

*Checkup ID:* 2526-001 | *Patient ID:* 2526 | *Gander:* Male | *Age:* 43
*Patient Record:* Follow-up consultation. Diagnosis: Dental arch defect. Periodontitis.
*Chief Complaint:* The patient reports mobility of porcelain crowns on lower anterior teeth for several months.
*Present Illness:* The patient has experienced mobility of porcelain crowns on lower anterior teeth for several months, affecting eating, and seeks restoration.
*Past Medical Record:* Generally healthy with no significant medical history.
*Examination:* Teeth 31, 32, 41, 42 have porcelain crown restorations. Teeth 41 and 31 are missing, with a splinted crown exhibiting grade III mobility. Teeth 32 and 42 have deep periodontal pockets, negative percussion and probing, mild gingival swelling, and bleeding on probing. No other abnormalities. Tooth 43 is missing with gingival swelling.
*X-Rays Examination:* Teeth 32 and 43 show alveolar bone width of 5mm and height of 12mm.
*Diagnosis:* Dental arch defect in teeth 31, 41, 43. Periodontitis in teeth 32, 42.
*Treatment Plan:* Extraction of teeth 32 and 42. Simultaneous lateral bone grafting and implantation at sites 32 and 43. Planned splinted crown restoration.
*Management:* For teeth 32 and 43, the patient was fully informed and consented to treatment, signing the informed consent for oral implant restoration. Routine disinfection and draping were performed, with local infiltration anesthesia using 4% Articaine. A horizontal incision was made at the alveolar ridge, teeth 32 and 42 were extracted, a full-thickness flap was elevated, and implant sites were prepared to the predetermined depth. Bego implants (3.75x10mm) were placed in positions 32 and 43, with cover screws inserted. A tenting screw was placed on the labial side, bone graft material was applied, the flap was sutured with tension-free closure, and post-operative radiographs were taken.
*Medical Instructions:* Suture removal in 1-2 weeks.
*Remarks:*

Fig. 2 A randomly selected sample from the dataset, showing intraoral photographs, radiographs, and the corresponding translated diagnostic report.

## 2.3. Data Privacy and Ethics

Throughout the data collection process, patient privacy and ethical compliance were prioritized. All patients provided written informed consent permitting the use of their anonymized clinical data for research and dataset development, and for minors, informed consent was obtained from their legal guardians. The study protocol for data collection was reviewed and approved by the local institutional ethics committee, ensuring adherence to applicable guidelines. All collected data were de-identified at the point of export, with personal identifiers removed and replaced by random numeric codes. Each patient received a unique identifier unrelated to their hospital ID, and each visit was assigned a visit number. Direct identifiers such as names, ID numbers, and contact details were eliminated, ensuring compliance with health data privacy regulations and making re-identification impossible. Developed under the supervision of the local ethics board, the dataset fully complies with recognized ethical and privacy standards for clinical research.

## 2.4. Data Annotation

To validate the dataset and evaluate AI dental assistants, the COde dataset was annotated for two benchmarking tasks: classification and generation. The classification benchmark measures model performance in detecting dental anomalies, while the generation benchmark evaluates the ability to produce detailed diagnostic reports from multimodal inputs. The generation task does not require explicit labeling, as models are prompted to emulate a dentist's report. For classification, disease labels were extracted directly from textual diagnostic reports written by dentists during routine checkups. Over 120 unique oro-dental conditions were identified across 8,000 clinical records, and labels were preserved in their original form to maintain consistency with the source diagnosis. A new column, *Diseases*, was added to the dataset to store these anomalies (e.g., caries, pulpitis).

To ensure accuracy, the auto-extracted labels were verified through a web-based annotation app specifically developed for this task. This app presented annotators with the full patient record, including intraoral photos, radiographs, and textual diagnostic reports, allowing holistic review of each case. Access was restricted to licensed dental practitioners familiar with clinical terminology and diagnostic procedures. Their task was not to re-diagnose but to confirm whether the extracted anomalies matched the clinician's report, ensuring textual consistency and correctness. This rigorous annotation process produced reliable benchmarks. A summary of dataset statistics is shown in Fig. 3

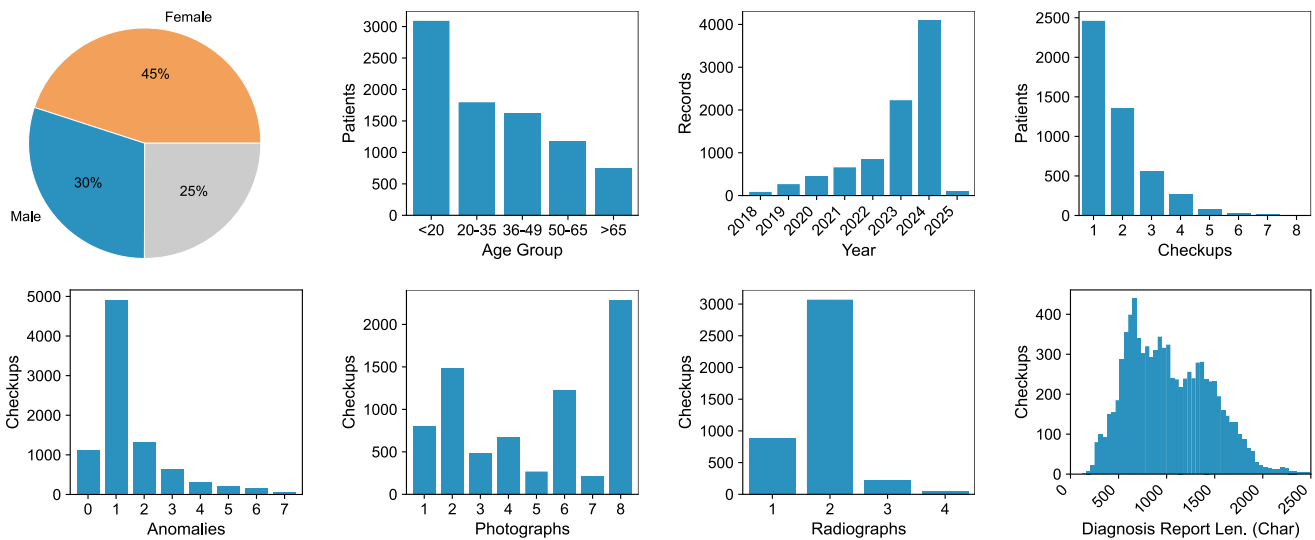

Fig. 3 Basic dataset statistics, including patient gender distribution, age, checkup year, number of checkups, anomalies count, associated photographs and radiographs, and textual report length.

## 3 Data Records

The dataset is organized as a multimodal dataset that links each patient's images with their textual records. All images are stored in two subfolders: *Photographs* and *Radiographs*. Each image filename consists of the *checkup_id* followed by a sequential index, which differentiates multiple files from the same visit. The textual data is provided in a CSV file, where each row corresponds to a single patient visit and contains the filenames of the associated images. Additionally, an alternate version of the dataset is provided in the ShareGPT format [19], a widely used data representation technique for VLMs training. In this format, each visit is represented as a JSON conversation combining textual and visual data. The directory structure of the dataset is illustrated in Fig. 4.

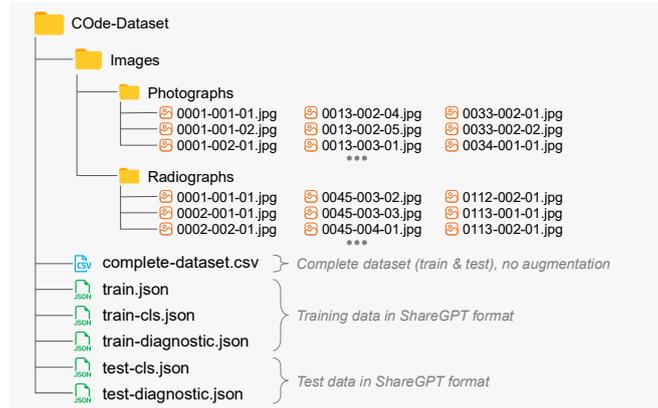

Fig. 4 Directory structure of the dataset, showing the organization of patient images along with the corresponding textual records provided in CSV and JSON formats.

## 4 Technical Validation

For the technical validation of our dataset's utility and effectiveness, we fine-tuned a state-of-the-art VLM and evaluated its performance before and after fine-tuning. Our hypothesis was that a model fine-tuned on the dataset would perform better than the base (non-fine-tuned) model on dentistry-related tasks. The base model served as the baseline, and any performance improvements in the fine-tuned model can be attributed to knowledge gained from the dataset. The models selected for the experiments were Qwen-VL-3B and Qwen-VL-7B, and GPT-4o.

### 4.1. Benchmarks

We designed two evaluation tasks: (1) dental anomaly classification task, and (2) diagnostic report generation task. A test set of 600 records was separated from the dataset to serve as the evaluation benchmark, while the remaining data were used for fine-tuning. The test set selection was not purely random; instead, we chose samples with the most frequent anomalies, as shown in Fig. 5, ensuring that those anomalies had sufficient representation in the training set.

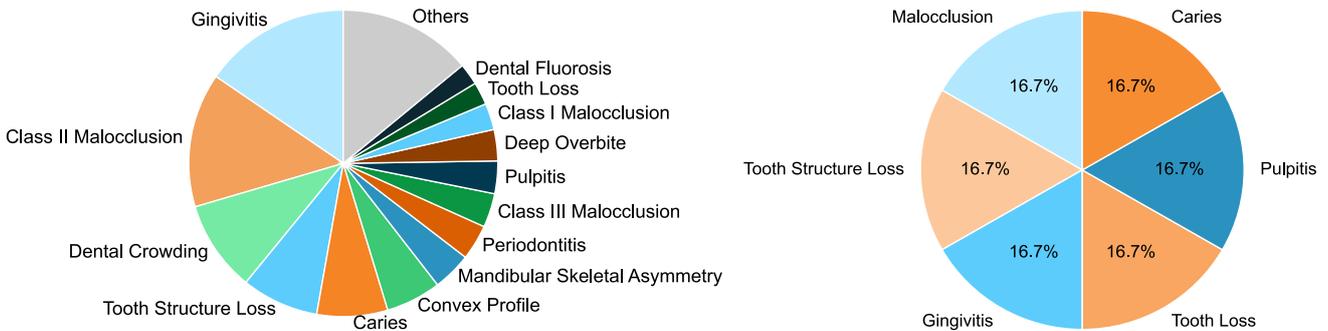

Fig. 5 Distribution of oro-dental anomalies: (A) In the entire dataset (B) In the test set only

**Classification Benchmark:** In this benchmark, the input fields are: *Age, Gender, Chief Complaint, Radiographs,* and *Photographs*, as shown in Table 2. Given these records, the model was asked to identify the anomaly, choosing from six categories: *Caries, Gingivitis, Malocclusion, Pulpitis, Tooth Loss,* or *Tooth Structure Loss*. We used standard classification metrics for evaluation: precision, recall, F1-score, and accuracy. The ground-truth labels were the anomalies extracted during the data annotation phase and manually verified by human experts.

**Generation Benchmark:** The input fields for this task are *Age, Gender, Chief Complaint, Past Medical History, Radiographs,* and *Photographs*. The model was prompted to generate a complete diagnostic report for the patient, emulating the detailed report a dental professional would write. The expected output report was structured to include these sections: *Patient Record, Clinical Examination, Radiographic Examination, Diagnosis, Treatment Plan, Management, Medical Instructions,* and *Remarks*. Evaluation metrics for the generation benchmark are BLEU, METEOR, and Cosine Similarity. BLEU and METEOR assess the degree of overlap between the generated text and the ground-truth report at the lexical and semantic levels, respectively. Cosine Similarity, on the other hand, measures the closeness of meaning and style by comparing text embeddings of the generated and reference reports. To mitigate potential bias from any single embedding model, we computed embeddings using two different LLMs, Gemma-2B [20] and Llama-3.2-1B [21], and averaged the two cosine similarity scores.

| Benchmark | Samples | Input Fields | Output Fields |
|---|---|---|---|
| Classification | 600 | Age, Gender, Chief Complaint, Radiographs, Photographs | Oro-dental anomaly |
| Generation | 600 | Age, Gender, Chief Complaint, Past Medical History, Radiographs, Photographs | Patient Record, Examination, Radiographs Examination, Diagnosis, Treatment Plan, Management, Medical Instructions, Remarks |

Table 2. Overview of classification and generation benchmarks, detailing input fields and expected outputs.

### 4.2. Baseline

Without any domain-specific fine-tuning, we first established baseline performance levels on our tasks using the base models. We considered two inference strategies for the base models: a zero-shot approach and a few-shot approach.

**Zero-Shot Prompting:** In the zero-shot scenario, the base model was given only a textual instruction (prompt) along with the input fields, without any example cases, and was asked to produce the desired output. For the classification task, for instance, we presented the model with the fields listed in Table 2 and prompted it to identify the oro-dental anomaly from the six categories. The model responded with a classification label. We applied a similar zero-shot prompting approach for the generation task, where the model was prompted to produce a detailed diagnostic report given the inputs, with no examples provided. The model generated a diagnostic report solely based on its pre-trained knowledge and understanding of the prompt.

**Few-Shot Prompting:** For the report generation task, the prompt in the few-shot setting included 3 examples of patient checkups, each with input fields and the corresponding ideal diagnostic report. This context was intended to guide the model to produce an output in the specific structured format and level of detail that matched the ground-truth reports. By contrast, we did not use a few-shot prompt for the classification task because the classification output is straightforward, and we observed that examples were not necessary for the model to understand the task.

### 4.3. Models Training (Finetuning)

Among the three models we evaluated, we fine-tuned the open-source Qwen-VL-3B and Qwen-VL-7B. The closed-source GPT-4o was not fine-tuned due to resource constraints and was therefore assessed only in zero-shot and few-shot settings. For training, we used the training split of the dataset released in ShareGPT format. Training ran on NVIDIA A800 GPUs: two cards for the 3B model and four for the 7B model. We adopted parameter-efficient LoRA [22] adapters applied to all transformer blocks, enabling stable updates with limited memory overhead. For Qwen-3B we used a batch size of 2; for Qwen-7B the batch size was 3. Both models were trained for three epochs with a cosine learning-rate schedule. Checkpoints with the best validation performance were retained for evaluation.

### 4.4. Results and Analysis

The classification results, summarized in Table 3, show a clear improvement in performance after fine-tuning on our dataset. The fine-tuned Qwen-7B achieved the best performance, with an accuracy of 78.92% and an F1-score of 79.39%, representing a substantial gain over all baseline models. In contrast, the lowest performance was observed with Qwen-3B in the zero-shot setting, which achieved 48.97% accuracy and an F1-score of 49.93%. Among the baselines, GPT-4o was the strongest, reaching about 55.83% accuracy. The confusion matrices for each model's predictions, shown in Fig. 6, further illustrate these outcomes: fine-tuned models produce much higher values along the main diagonal. The results of the diagnostic report generation task are presented in Fig. 7, where the evaluation metrics are BLEU, METEOR and cosine similarity between the generated report and the ground-truth report. Once again, the fine-tuned Qwen-7B delivered the best performance, achieving an average similarity score of 71.46%, indicating that its outputs were extremely close to dentist-written reports in both content coverage and wording. In comparison, the baseline models scored lower on this metric. Notably, GPT-4o with few-shot prompting produced the most faithful reports among the baselines. Across both benchmarks, the models fine-tuned on our dental dataset significantly outperformed their baseline counterparts. The fine-tuned Qwen-7B model in particular exhibited the highest efficacy on both tasks. These results validate the utility of our dataset, as the VLMs acquired specialized knowledge that enabled more accurate anomaly classification and human-like report generation. This evaluation confirms that our dataset is a valuable resource for advancing AI-driven dentistry.

| Model | Accuracy (%) | Precision (%) | Recall (%) | F1-Score (%) |
|---|---|---|---|---|
| Zero-Shot Qwen-3B | 48.97 | 59.35 | 48.97 | 49.93 |
| Zero-Shot Qwen-7B | 52.65 | 73.55 | 52.65 | 52.99 |
| Zero-Shot GPT-4o | 55.83 | 62.37 | 55.83 | 54.54 |
| SFT Qwen-3B | 74.90 | 79.97 | 74.90 | 76.19 |
| SFT Qwen-7B | 78.92 | 85.06 | 78.92 | 79.39 |

Table 3 Results of the models evaluated on the classification benchmark.

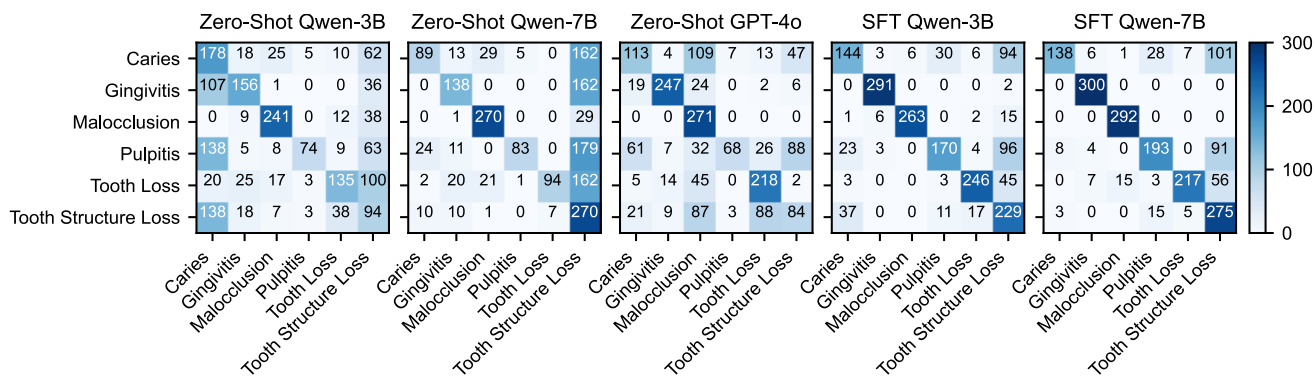

Fig. 6 Confusion matrices of classification predictions across models, with the main diagonal showing correct predictions and off-diagonal values representing incorrect predictions

## 5 Usage Notes

The dataset is publicly accessible on Hugging Face [23]. The textual part of the dataset is provided in both CSV and JSON formats. We have released the dataset in its original form, without QA queries or augmentation, allowing researchers to apply their own augmentation techniques and QA strategies. In addition, we provide a ready-to-use version in ShareGPT format, which includes two types of augmentation: (i) randomization of the image order, and (ii) random assignment of different QA pairs to each record. The ShareGPT dataset can be seamlessly integrated into machine learning pipelines with a single line of code using the HuggingFace Datasets library. Furthermore, training and test splits are provided as individual files for each benchmark, making it straightforward to train or evaluate models on specific tasks.

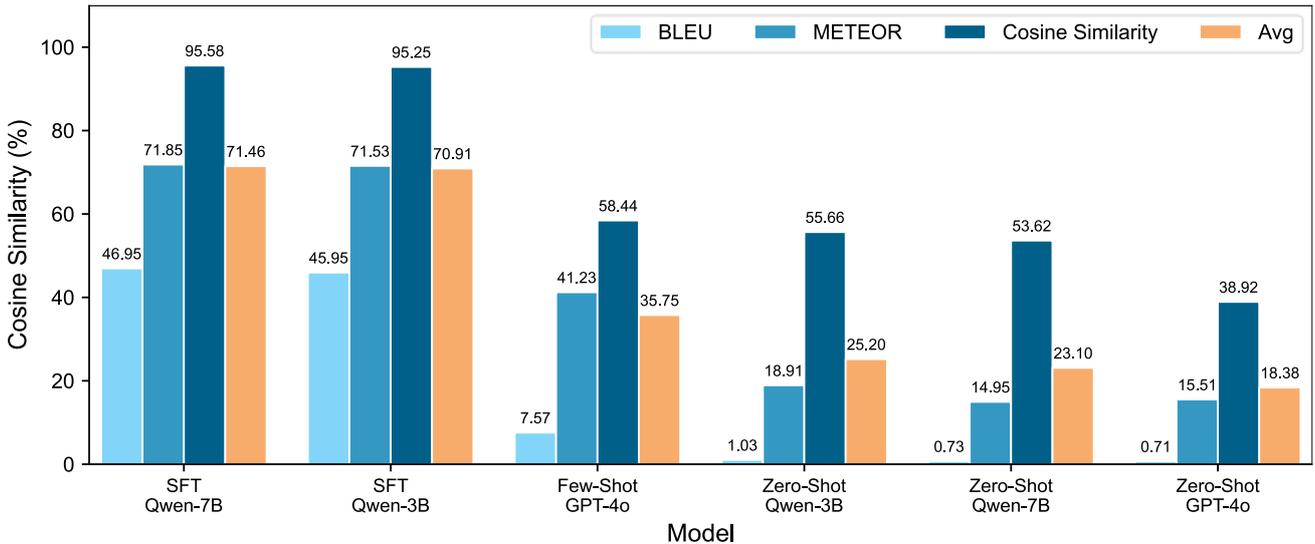

Fig. 7 Report generation benchmark results evaluated using BLEU, METEOR, and Cosine Similarity

## 6 Data Availability

The COde dataset is publicly available on Hugging Face, provided in compressed ZIP format, at the following link: https://huggingface.co/datasets/zirak-ai/COde.

## 7 Code Availability

Custom Python scripts were developed for anonymizing and preprocessing patient data, as well as for training and inference of VLMs. These codes were created solely for internal use and are not publicly released.